\documentclass[journal]{IEEEtran}
\usepackage{graphicx,amssymb,mathrsfs,amsmath,array,color,hhline}
\usepackage{subfigure}
\usepackage{multirow}
\usepackage{booktabs}
\usepackage{algorithm}
\usepackage{algpseudocode}
\usepackage{cite}
\usepackage[colorlinks,urlcolor=blue,driverfallback=dvipdfm]{hyperref}

\begin{document}
\title{Multimodal Collaboration Networks for Geospatial Vehicle Detection in Dense, Occluded, and Large-Scale Events }

\author{Xin Wu,~\IEEEmembership{Senior Member,~IEEE,}
        Zhanchao Huang, ~\IEEEmembership{Member,~IEEE,}
        Li Wang,~\IEEEmembership{Senior Member,~IEEE,}
        Jocelyn Chanussot,~\IEEEmembership{Fellow,~IEEE,}
        and Jiaojiao Tian,~\IEEEmembership{Senior Member,~IEEE}
        
\thanks{This work was supported by the National Natural Science Foundation of China under Grant 62101045, the Natural Science Foundation of Beijing Municipality under Grant L222041, the MIAI@Grenoble Alpes (ANR-19-P3IA-0003), and the Scientific Research Startup Fundation of Fuzhou University Grant 511271. (\emph{Corresponding author: Zhanchao Huang})}
\thanks{X. Wu and L. Wang is with the School of Computer Science (National Pilot Software Engineering School), Beijing University of Posts and Telecommunications, Beijing 100876, China. (e-mail: 040251522wuxin@163.com)}
\thanks{Z. Huang is with the Key Laboratory of Spatial Data Mining and Information Sharing of the Ministry of Education, The Academy of Digital China, Fuzhou University, Fuzhou 350108, China; and with the National \& Local Joint Engineering Research Center of Satellite Geospatial Information Technology, Fuzhou University, Fuzhou 350108, China. (e-mail: huangc@fzu.edu.cn).}
\thanks{J. Chanussot is with the Univ. Grenoble Alpes, CNRS, Grenoble INP, GIPSA-Lab, 38000 Grenoble, France, also with the Aerospace Information Research Institute, Chinese Academy of Sciences, Beijing 100094, China. (e-mail: jocelyn.chanussot@inria.fr)}
\thanks{J. Tian is with the Remote Sensing Technology Institute, German Aerospace Center, 82205 We{\ss}ling, Germany. (e-mail: jiaojiao.tian@dlr.de).}
}

\markboth{IEEE Transactions on Geoscience and Remote Sensing, ~Vol.~62, No.~5616812,~2024}
{Shell \MakeLowercase{\textit{et al.}}: }
\maketitle

\begin{abstract}
In large-scale disaster events, the planning of optimal rescue routes depends on the object detection ability at the disaster scene, with one of the main challenges being the presence of dense and occluded objects. Existing methods, which are typically based on the RGB modality, struggle to distinguish targets with similar colors and textures in crowded environments and are unable to identify obscured objects. To this end, we first construct two multimodal dense and occlusion vehicle detection datasets for large-scale events, utilizing RGB and height map modalities. Based on these datasets, we propose a multimodal collaboration network for dense and occluded vehicle detection, MuDet for short. MuDet hierarchically enhances the completeness of discriminable information within and across modalities and differentiates between simple and complex samples. MuDet includes three main modules: Unimodal Feature Hierarchical Enhancement (Uni-Enh), Multimodal Cross Learning (Mul-Lea), and Hard-easy Discriminative (He-Dis) Pattern. Uni-Enh and Mul-Lea enhance the features within each modality and facilitate the cross-integration of features from two heterogeneous modalities. He-Dis effectively separates densely occluded vehicle targets with significant intra-class differences and minimal inter-class differences by defining and thresholding confidence values, thereby suppressing the complex background. Experimental results on two re-labeled multimodal benchmark datasets, the 4K-SAI-LCS dataset, and the ISPRS Potsdam dataset, demonstrate the robustness and generalization of the MuDet. The codes of this work are available openly at \url{https://github.com/Shank2358/MuDet}.

\end{abstract}

\graphicspath{{figures/}}


\begin{figure}[!t]
    \begin{center}
        \includegraphics[width=0.5\textwidth]{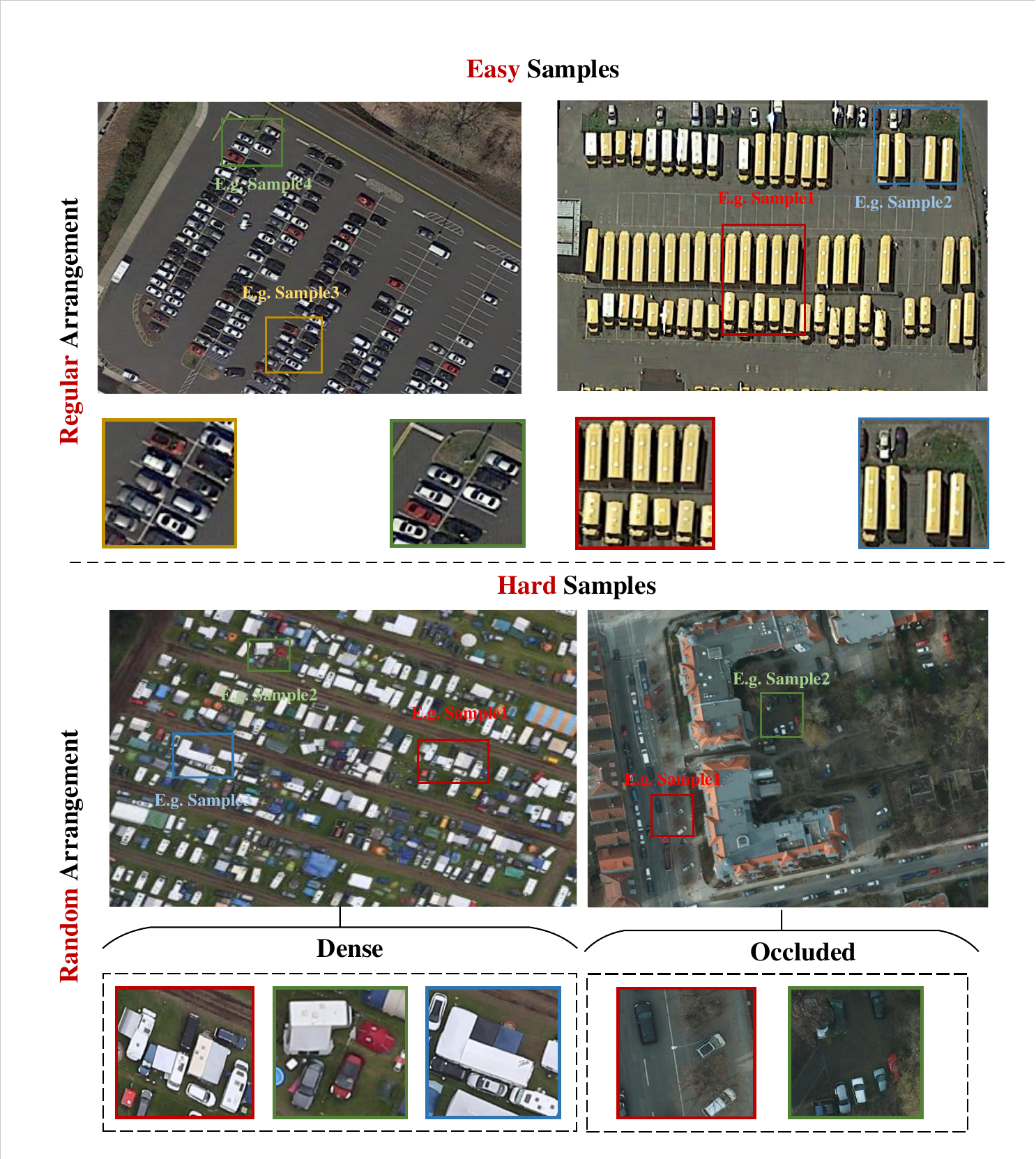}
        \caption{A visual example for the separable and non-separable objects in large-scale disaster events.}
        \label{fig:demo}
    \end{center}
\end{figure}

\section{Introduction}
\IEEEPARstart{R}{emote} sensing (RS) imagery has long been utilized across various fields of large-scale disaster events, including early warning and damage evaluation. With the rapid development of domestic and foreign satellites and aircraft, as well as the popularity of unmanned aerial vehicles (UAVs) \cite{wu2022deep}, the spatial and spectral resolutions of RS data have been continuously improved \cite{hong2024spectralgpt}, providing basic data assurance for applications in large-scale disaster events. Object detection \cite{weber2021artificial,wu2022uiu,8654203} in RS images is the technical basis for risk assessment and rescue. However, research focusing on object detection, e.g., vehicle detection, in large-scale events, which involves challenges such as density, occlusion, and even distortion, remains relatively limited.

In remote sensing (RS) object detection, especially when utilizing deep learning (DL) models, high-quality labeling is crucial for defining precise object boundaries and categories. This accuracy is vital for the model's ability to learn and recognize distinctive features of each object class. The Northwestern Polytechnical University Very-High-Resolution dataset (NWPU VHR-10)\cite{cheng2016learning}, the INDIA aerial picture dataset, UCAS-AOD \cite{zhu2015orientation}, the Remote Sensing Object Detection (RSOD) dataset \cite{long2017accurate}, and the Dataset for Object deTection (DOTA)\cite{xia2018dota}, are representative examples of publicly available object detection datasets. The majority of these datasets are sourced from Google Maps with RGB modality, with the vehicle targets predominantly located in parking lots, roadside areas, residential zones, and other relevant scenes. All datasets are in RGB modality, with minor variations among the image scenes. There is a significant difference between different vehicle target classes, yet there is a high degree of similarity within each class. Additionally, vehicles in these scenes adhere to predefined parking rules and regulations. A selection of annotated vehicle samples is presented above the dashed line in Fig. \ref{fig:demo}.

\begin{figure*}[!t]
    \begin{center}
        \includegraphics[width=1.0\textwidth]{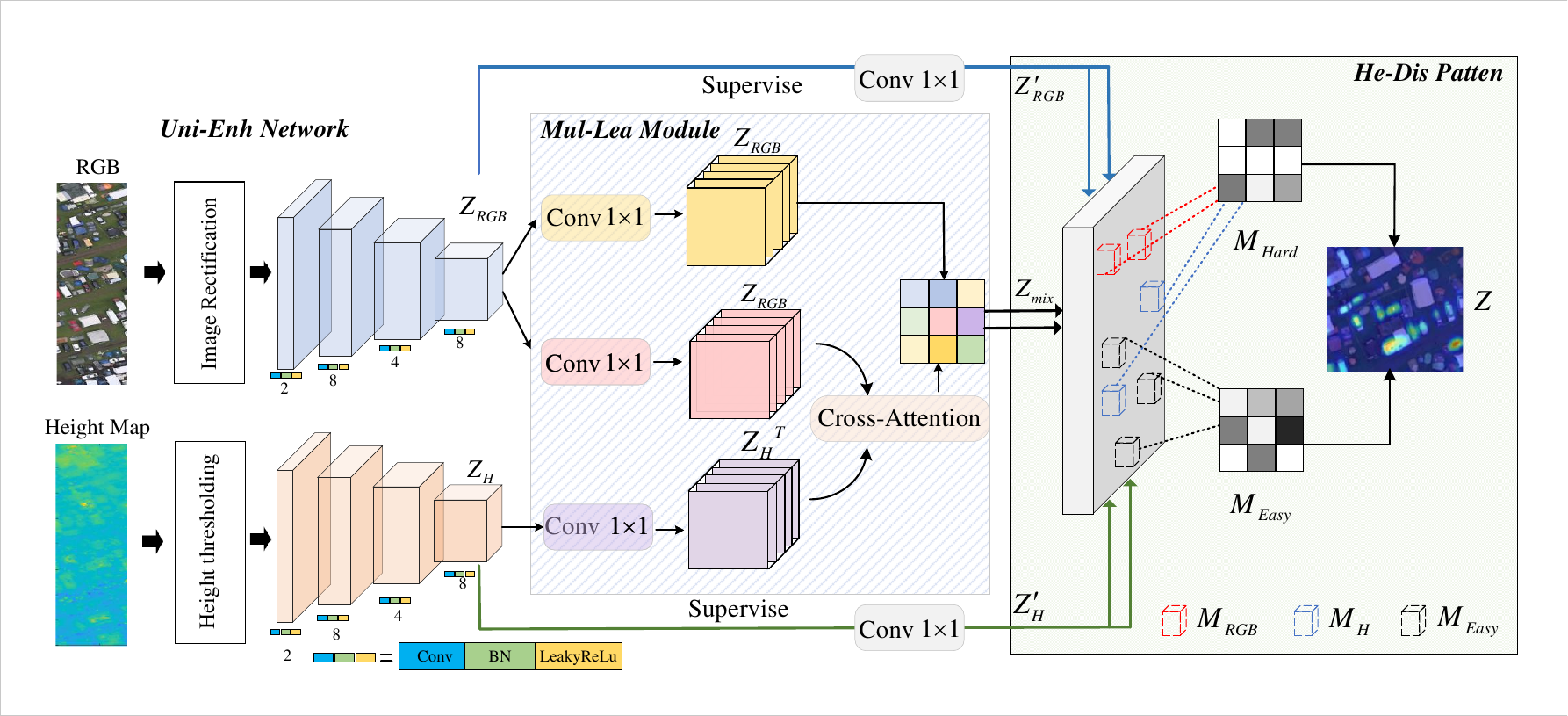}
        \caption{The flowchart of the proposed MuDet for dense and occluded vehicle detection in multimodal RS images.}
        \label{fig:outline}
    \end{center}
\end{figure*}

Currently, there are numerous methods for detecting the above vehicles. For instance, Ref. \cite{wang2017fast} utilized adversarial learning to create vehicle images to diversify the dataset and enhance detection accuracy. To concentrate on regions of interest while minimizing the impact of occlusions, Zhang et al. \cite{rs12111760} developed a triple-head network incorporating regional attention. Zhu et al. \cite{zhu2021spatial} proposed a Hard Samples Metric Learning (HSML) strategy aimed at reducing intra-class variance and lowering the rate of false detections. Meanwhile, Huang et al. \cite{huang2022general} developed an Object-Adaptation Label Assignment (OLA) method that adapts neural network learning to the specific characteristics of different objects, indirectly addressing the challenge of densely packed boats and vehicles. Unfortunately, RGB modality data alone falls short of distinguishing between objects that are densely packed or occluded within the small intra-class distance. This limitation is particularly evident at large-scale events, for example, where recreational vehicles (RVs), vehicles with tents mounted on their roofs, and flat-topped rectangular tents are densely parked in a limited area. This scenario is depicted in the example images located below the dotted line in Fig. \ref{fig:demo}.

Multimodal data \cite{hong2023decoupled}, which integrates information from various sensors or sources, such as visible light (RGB), infrared, light detection and ranging (LIDAR), synthetic aperture radar (SAR), and optical photographic measuring, significantly enhances the ability to detect and differentiate vehicles. This integration leverages the strengths of each modality to overcome the limitations inherent in any single data type, particularly in challenging conditions such as denseness, occlusion, or similar appearance among objects. Sharma et al. \cite{sharma2020yolors} utilized mid-level feature fusion to integrate data from visible and infrared (IR) modalities. Sumbul et al. \cite{sumbul2019multisource} developed a unified object detection framework that integrates feature representations and attention mechanisms from both visible and LIDAR data. A significant challenge in multimodal object detection is crafting effective fusion strategies. In response, Hong \textit{et al.} \cite{hong2023cross} prompted a promising research problem, i.e., cross-region or cross-city land cover classification, and proposed a novel multimodal deep learning method, called high-resolution domain adaptation networks. However, current multimodal datasets and their associated methodologies mainly concentrate on object deformation, such as variations in scale and orientation, while often neglecting issues like irregular parking and obstructions, e.g., tents or branches. This oversight leads to significant challenges in detecting vehicles amidst dense occlusion at large-scale events. 

To this end, we first construct and label two Multimodal Vehicle Detection (MVD) datasets at large-scale events, incorporating both RGB and height map modalities. These datasets are characterized by densely packed vehicles, occlusions, and instances of partial deformation. RGB images provide color and texture information about objects, aiding in the identification of vehicular surface characteristics. Height maps offer elevation data for objects, enabling the differentiation of objects with similar colors and textures in crowded environments by their distinct heights. Furthermore, height maps can also infer the presence of occluded objects based on their height differences. Then, we propose a multimodal collaboration network for dense and occluded vehicles. Specifically, we design a Unimodal Feature Hierarchical Enhancement (Uni-Enh) network and a Multimodal Cross Learning (Mul-Lea) strategy to enhance the distinct features of each modality and enrich the feature representation of vehicles. Following it, a Hard-Easy Discriminative (He-Dis) pattern is designed to enhance the discriminability between hard and easy objects and to minimize the impact of complex background interference. The contributions of this paper are summarized as follows:

\begin{itemize}
    \item A multi-modal vehicle detection dataset is constructed and labeled, specifically targeting vehicles in dense and occluded scenarios in large-scale events. These vehicles are categorized as ``hard vehicles'' due to the complexity of their detection conditions.
    
    \item A Multimodal Collaboration Network (MuDet) is proposed to detect dense and occluded vehicles in large-scale events. By integrating RGB and height map data, it enhances features within each modality and improves the completeness of feature fusion across modalities. MuDet significantly enhances the discriminability and separability of multimodal features.
    
    \item A unimodal feature hierarchical enhancement (Uni-Ehn) network and a multimodal cross learning (Mul-Lea) strategy are designed to enhance the distinct features of each modality and enrich the distinguishing features of vehicles. A hard-easy discriminative pattern (He\_Dis) module is designed to balance hard-easy object discriminability and suppress interference from complex backgrounds on objects. 
       
   \item We evaluate the detection performance of the proposed MuDet on two new multimodal vehicle detection datasets, namely the 4K-SAI-LCS dataset and the ISPRS Potsdam dataset, demonstrating substantial improvements over various existing methods. The codes and datasets will be available for the sake of reproducibility and for developing the research direction of multimodal RSOD.
   
\end{itemize}

This paper is organized as follows: Section \ref{sec:Me} introduces the proposed MuDet framework in detail, including the Uni-Ehn network, Mul-Lea strategy, He-Dis modules, and its loss function. Quantitative experiments and visual analysis of the proposed MuDet on two newly annotated multimodal datasets, the 4K-SAI-LCS dataset and the ISPRS Potsdam dataset, are discussed in Section \ref{sec:Exp}. Section \ref{sec:clu} summarizes and offers prospects for the proposed MuDet framework.

\section{Proposed MuDet Framework}\label{sec:Me}

In this section, we provide a detailed description of the proposed MuDet for dense and occluded vehicle detection. Fig. \ref{fig:outline} illustrates the network architecture of MuDet. In detail, we first introduce an unimodal feature learning and enhancement (Uni-Enh) network designed to amplify the distinctive features of each stream, aiming to capture intramodal relationships more precisely. Then, the generated feature is sent to the multimodal cross-learning (Mul-Lea) module to interactively learn features between two heterogeneous modalities and improve the completeness of information fusion across these modalities. To further enhance the detection of vehicles in dense and occluded conditions, we developed a hard-easy discriminative (HE-Dis) pattern that differentiates vehicles across varying levels of density and occlusion. 

\subsection{Unimodal Feature Hierarchical Enhancement (Uni-Enh)}\label{sec:B}
Convolutional Neural Networks (CNNs) feature a unique architecture of local weight sharing, significantly benefiting image processing and other areas by enabling the extraction of highly discriminative object features from input images. Currently, CNNs are the widely used method in the field of remote sensing \cite{li2020object,li2023lrr}. Unimodal feature hierarchical enhancement (Uni-Enh) involves dual-stream feature learning via CNNs and features hierarchical enhancement of each stram to more effectively capture intramodal relationships.

Firstly, we introduce a dual-stream CNN-based network for feature learning. Each stream in the network is composed of several CNN blocks, with each block consisting of a $3\times 3$ convolutional layer, Batch Normalization (BN), and Leaky ReLU activation. We define ${\boldsymbol{X}} \in {\mathcal{R}^{{d} \times M\times N}}$ as the CNN features. To distinguish, $\mathbf{X}_{RGB} \in \mathcal{R}^{{d_{RGB}} \times M\times N}$ and $\mathbf{X}_H \in \mathcal{R}^{{d_{H}} \times M\times N}$ are used to represent the feature maps of the RGB image and the height map (H), respectively. ${d_{RGB}}$ and ${d_{H}}$ represent the number of RGB stream channels and the number of height map stream channels, respectively. $M\times N$ denotes the feature map size. The output of the $l$-th layer of MuDet is denoted as $\boldsymbol{X}^{(l)}$.

\begin{equation}
\label{eq1}
    \boldsymbol{X}^{(l)} =f( \boldsymbol{W}^{(l)}\boldsymbol{X}^{(l-1)} + \boldsymbol{b}^{(l)}), l = 1,2, \cdots ,N_s,
\end{equation}
where $\boldsymbol{X}^{(l)}$ denotes the feature maps of the $l$-th layer. $f$ is a nonlinear activation function. $N$ indicates the layer number of CNN. $\boldsymbol{W}^{(l)}$ and ${\boldsymbol{b}^{(l)}}$ are the learned weights and biases of the $l$-th layer, respectively. 

Then, we develop a hierarchical enhancement strategy to amplify the distinctive features of each stream and integrate the outcomes into a cross-attention mechanism, thereby aiming to more precisely capture intramodal relationships.

For the RGB stream, we measure the grayscale values of the RGB image $\boldsymbol{X}_{RGB}^{(0)}$ and employ gamma transformation with diverse coefficients to refine details across both low and high grayscale spectrums. Thus, the input of the RGB stream is

\begin{equation}
\label{CA}
  \boldsymbol{\hat{X}}_{RGB}^{(0)}= \gamma({A{\boldsymbol{X}_{RGB}^{(0)}}}), 
\end{equation}
where $A$ is a constant and $\gamma$ represents the gamma transformation function.

For the height map stream ${\boldsymbol{X}_H^{(0)}}$, we employ grayscale slicing to emphasize the height information of foreground objects while masking the height values of background objects, leveraging expert prior knowledge.
\begin{equation}
 \label{CA}
    {X_H^{(0)}}(i,j) = \left\{ \begin{array}{l}
    \begin{array}{*{20}{c}}
    {\begin{array}{*{20}{c}}
    H_1&{}&{}
    \end{array}}&{{C_{\min }} \le }
    \end{array}{X_H^{(0)}}(i,j) \le {I_0}\\
    \begin{array}{*{20}{c}}
    {{X_H^{(0)}}(i,j)}&{{I_0}}
    \end{array} \le {X_H^{(0)}}(i,j) \le {I_1}\\
    \begin{array}{*{20}{c}}
    {\begin{array}{*{20}{c}}
    H_2&{}&{}
    \end{array}}&{{I_1} \le }
    \end{array}{X_H^{(0)}}(i,j) \le {C_{\max }}
    \end{array} \right.
\end{equation}
where $H_1$ and $H_2$ are the experiential thresholds for background objects. $\boldsymbol{X}_{H}^{(0)}{(i,j)}$ represents the height map value at a specific position $(i,j)$. The values ${I_0, I_1}$ denote distinct slicing thresholds. The constants $C_{\min}$ and $C_{\max}$ signify the minimum and maximum height values, respectively.

\subsection{Multimodal Cross Learning (Mul-Lea)} \label{sec:B}

The concept of cross-attention, as first introduced in the transformer architecture for language processing due to its potent semantic feature extraction and long-range feature capture capabilities \cite{yao2023extended}. It asymmetrically combines two independent sequences of embeddings, each with the same dimensions. Here, the two sequences correspond to the features of the two modalities. Given the feature map $\boldsymbol{Z}_{RGB} = \boldsymbol{X}_{RGB}^{(N_{RGB})}$ and $\boldsymbol{Z}_{H} = \boldsymbol{X}_{H}^{(N_{H})}$ of two modalities, the cross-attention mechanism is defined as follows: 
\begin{figure}[!t]
    \begin{center}
        \includegraphics[width=0.5\textwidth]{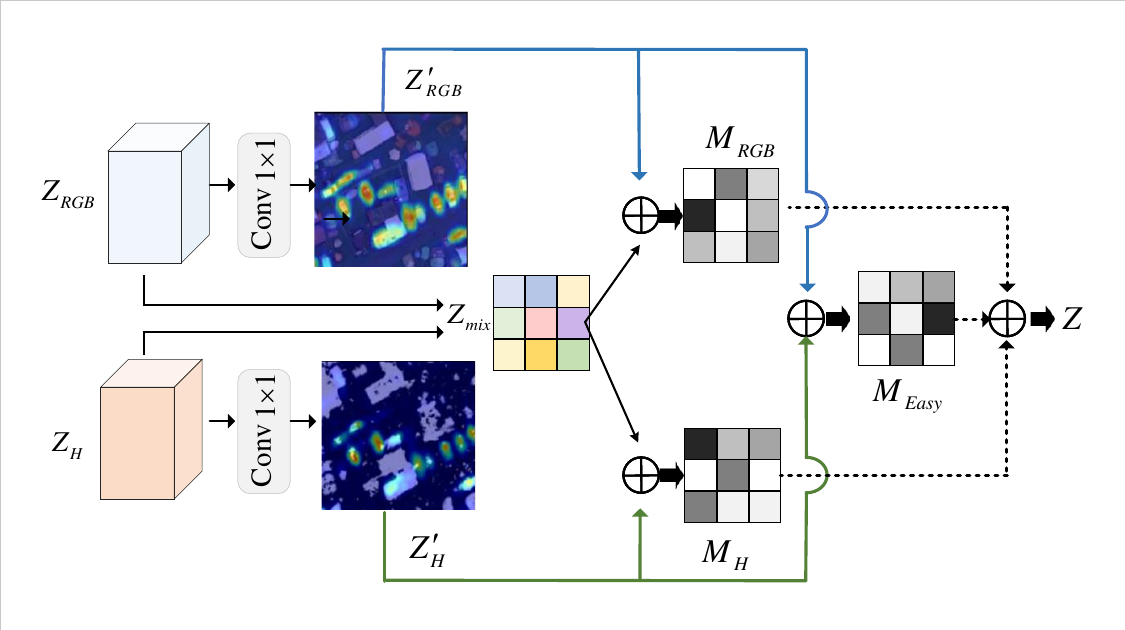}
        \caption{An illustration of the hard-easy discriminative pattern.}
        \label{fig:hard}
    \end{center}
\end{figure}
\begin{equation}
\label{CA}
    \begin{aligned}
    \mathbf{Z}_{mix} &= Attention(\boldsymbol{Z}_{RGB},\boldsymbol{Z}_{H}) \\ &= \boldsymbol{Z}_{RGB}\boldsymbol{Z}_{H}^T\boldsymbol{Z}_{RGB}\\
    &=Softmax(( {\frac{{\boldsymbol{Q}{\boldsymbol{K}^T}}}{{\sqrt d }}} ) \boldsymbol{V}), 
    \end{aligned}
\end{equation}
where $\boldsymbol{Q} = g_1(\boldsymbol{Z}_{RGB})$, $\boldsymbol{K} = g_2(\boldsymbol{Z}_{H})$, and $\boldsymbol{V} = g_3(\boldsymbol{Z}_{RGB})$, with $g_1$, $g_2$, and $g_3$ being $1 \times 1$ convolutions. Thus, $\boldsymbol{Z}_{mix}$ is also represented as follows:
\begin{equation}
\mathbf{Z}_{mix} =Softmax(( {\frac{{g_1(\boldsymbol{Z}_{RGB}){g_2(\boldsymbol{Z}_{H})^T}}}{{\sqrt d }}} )g_3(\boldsymbol{Z}_{RGB})).
\end{equation}

Overall, the Uni-Enh block and the Mul-Lea block hierarchically enhance vehicle differentiation within each modality and interactively learn features between two heterogeneous modalities, respectively. They dynamically improve the completeness of information fusion both within and across modalities, resulting in enhanced multimodal feature discriminability and separability.

\subsection{Hard-easy Discriminative (He-Dis) Pattern} \label{sec:C}
To enhance the distinction and detection of vehicles in large-scale events, we designed a hard-easy discriminative pattern. This pattern begins by calculating the confidence value of features within each modality, followed by constructing and thresholding \textit{easy-to-predict} and \textit{hard-to-predict} masks to accurately detect vehicles. This pattern ensures precise supervision of each modality, facilitating a more effective vehicle location.
Fig. \ref{fig:hard} shows an illustration of the hard-easy discriminative pattern. More specifically, we define the confidence value as $\boldsymbol{Conf}$, 

\begin{equation}
\label{CA}
    \begin{aligned}
    \boldsymbol{Conf}_{RGB} &= Sigmoid({h_{RGB}(\boldsymbol{Z}_{RGB})}) \\ &= \frac{1}{{1 + {e^{ - h_{RGB}(\boldsymbol{Z}_{RGB})}}}}, 
    \end{aligned}
\end{equation},

\begin{equation}
\label{CA1}
    \begin{aligned}
    \boldsymbol{Conf}_{H} &= Sigmoid({h_{H}(\boldsymbol{Z}_{H})}) \\ &= \frac{1}{{1 + {e^{ - h_{H}(\boldsymbol{Z}_{H})}}}}, 
    \end{aligned}
\end{equation}
where $h_{RGB}(.)$ and $h_H(.)$ represent the $1\times 1$ convlutions.

To accurately differentiate between hard and easy vehicles, we define a threshold $\theta$. If the vehicle confidence predicted by both the RGB stream and height map stream exceeds the threshold $\theta$, the vehicle predicted at position $(i, j)$ is classified as an \textit{easy-to-predict} sample, and a mask $Mask_{easy}$ is given.

\begin{equation}
{\mathbf{M}_{easy}} = \left\{ \begin{array}{l}
\begin{array}{*{20}{c}}
1&{{Conf}_{RGB}}
\end{array}( {i,j} )> \theta,{{Conf}_{H}}( {i,j}) > \theta\\
\begin{array}{*{20}{c}}
0&{others}
\end{array}
\end{array} \right.
\end{equation}

If the object confidence predicted by either the RGB branch or the height map branch is greater than the threshold $\theta$, and the other is less than $\theta$, indicating that not both modal features can detect the object, then the object predicted at position $(i, j)$ is considered a \textit{hard-to-predict} sample. Thus, two masks $Mask_{RGB}$ and $Mask_{H}$ are given, 

\begin{equation}
{\mathbf{M}_{RGB}} = \left\{ \begin{array}{l}
\begin{array}{*{20}{c}}
1&{{Conf}_{RGB}}
\end{array}( {i,j} ) > \theta,{{Conf}_{H}}( {i,j} ) < \theta\\
\begin{array}{*{20}{c}}
0&{others}
\end{array}
\end{array} \right.
\end{equation}
\begin{equation}
{\mathbf{M}_H} = \left\{ \begin{array}{l}
\begin{array}{*{20}{c}}
1&{{Conf}_{RGB}}
\end{array}( {i,j} ) < \theta,{{Conf}_{H}}( {i,j}) > \theta\\
\begin{array}{*{20}{c}}
0&{others}
\end{array}
\end{array} \right.
\end{equation}

Finally, all detected vehicles can be formulated as follows: 
\begin{equation}
    \begin{aligned}
    \mathbf{Z} &= Mas{k_{easy}} \cdot ({\mathbf{Z}_{mix}} + {\mathbf{Z}_{RGB}} + {\mathbf{Z}_{H}})  \\
    &+ Mas{k_{RGB}} \cdot ({\mathbf{Z}_{mix}} + {\mathbf{Z}_{RGB}}) \cdot ( {2 - Con{f_{RGB}}} ) \\
    &+ Mas{k_H} \cdot ({\mathbf{Z}_{mix}} + {\mathbf{Z}_{H}}) \cdot ( {2 - Con{f_{H}}} ).
    \end{aligned}
\end{equation}

The hard-easy discriminability strategy streamlines the differentiation between hard and easy vehicles through soft thresholding with the hard-easy mask, thereby significantly improving the separation of dense and occluded vehicles.

\subsection{Loss Function} \label{sec:D}
In this section, we employ distinct loss functions to supervise hard and easy vehicles separately.

\begin{figure}[!t]
    \begin{center}
        \includegraphics[width=0.4\textwidth]{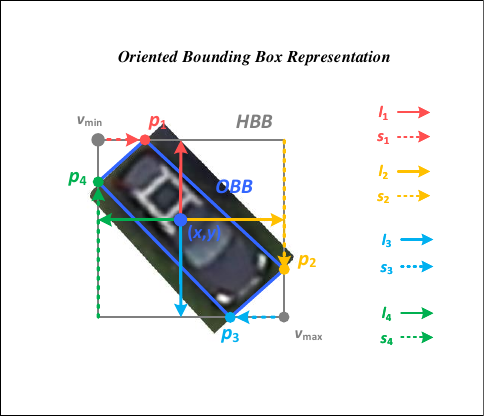}
        \caption{The details representation of the OBB.}
        \label{fig:obb}
    \end{center}
\end{figure}

\textit{For the easy-to-predict vehicles}, we employed two loss functions: an object classification loss, denoted as $L_{cls}$, and an Oriented Bounding Box (OBB) regression loss, represented as $L_{reg}$. Specifically, the classification loss is formulated using focal loss \cite{lin2017focal}, that is 
\begin{equation}
    {L^{cls}} =  - {(1 - p_t)^\gamma }\log {p_t},
\end{equation}
\begin{equation}
    {p_t} = \left\{ \begin{array}{l}
\begin{array}{*{20}{c}}
{\hat p}&{}&{\begin{array}{*{20}{c}}
{if}&{p = 1}
\end{array}}
\end{array}\\
\begin{array}{*{20}{c}}
{1 - \hat p}&{otherwise}
\end{array}
\end{array} \right.
\end{equation}
where $\hat{p}$ represents the predicted probability, and $p={0,1}$ denotes the true label. In alignment with the paper \cite{lin2017focal}, we set the hyperparameter $\gamma$ to 2.
 
To refine the regression results, the regression loss $L_{reg}$ employs the definition provided in Ref \cite{huang2022general}, utilizing OBB for more precise object localization.
\begin{equation}
\begin{array}{l}
    L^{reg} = 1 - IoU({\boldsymbol{l}_i}, {{\boldsymbol{\hat l}}_i})_{i = 1,2,3,4} \\
    + \sum\limits_{i = \mathbf{1}}^4 {{{( {\mathbf{s}_i} -  {\mathbf{\hat s}}_i}})^2}  + (r - \hat r)^2,
\end{array}
\end{equation}
where $\mathbf{l}$ and ${{\mathbf{\hat l}}}$ represent the distances from the sampling point to the horizontal bounding box (HBB) boundaries. $\mathbf{s}$ and $ {\mathbf{\hat s}}$ denote the distances between the HBB vertices and the OBB vertices. $r$ and ${\hat r}$ are the ratios of the HBB to the OBB in terms of area. $IoU()$ signifies the Intersection over Union (IoU) between two HBBs. Define the ground truth distances ${\boldsymbol{l}}$ is composed of ${l_1}$, ${l_2}$, ${l_3}$, ${l_4}$, and the predicted distances ${\boldsymbol{\hat l}}$ is composed of ${\hat l_1}$, ${\hat l_2}$, ${\hat l_3}$, ${\hat l_4}$. The area of ground truth HBB is $area{} = \left( {{l_1} + {l_3}} \right) \times \left( {{l_2} + {l_4}} \right)$ and the area of predicted HBB is ${\widehat {area}} = \left( {{{\hat l}_1} + \hat l} \right) \times \left( {{{\hat l}_2} + {{\hat l}_4}} \right)$. Then, the Overlapping area is represented as 
\begin{equation}
\begin{array}{l}
area^{overlap} = \left( {\min \left( {{l_1},{{\hat l}_1}} \right) + \min \left( {{l_3},{{\hat l}_3}} \right)} \right) \\
\times \left( {\min \left( {{l_2},{{\hat l}_2}} \right) + \min \left( {{l_4},{{\hat l}_4}} \right)} \right). 
\end{array}
\end{equation}
The area of the circumscribed HBB of the two HBBs above is represented as 
\begin{equation}
\begin{array}{l}
area^{circ} = \left( {\max \left( {{l_1},{{\hat l}_1}} \right) + \max \left( {{l_3},{{\hat l}_3}} \right)} \right) \\
\times \left( {\max \left( {{l_2},{{\hat l}_2}} \right) + \max \left( {{l_4},{{\hat l}_4}} \right)} \right).
\end{array}
\end{equation}
The area of the union region of the two HBBs above is represented as ${U_{x,y}} = are{a_{x,y}} + {\widehat {area}_{x,y}} - area_{x,y}^{overlap}$. Thus, 
\begin{equation}
\begin{array}{l}
IoU\left( {{\boldsymbol{l}_i},{\boldsymbol{\hat l}_i}} \right) =  \displaystyle{\frac{{area^{overlap}}}{U}}. 
\end{array}
\end{equation}

Thus, the loss of easy samples is
\begin{equation}
    L_{easy} = {\frac{1}{N}}\sum_{i=0}^N {(L^{reg}_i + {L^{cls}_i})},
\end{equation}
where $N$ represents the number of samples. Figure \ref{fig:obb} provides a detailed representation of the OBB.

\textit{For the hard vehicles}, the total loss is
\begin{equation}
    L_{hard} = {\frac{1}{N}}\sum_{i=0}^N {((Mask_{RGB}+Mask_{H})\dot (L^{reg}_i + {L^{cls}_i}))}.
\end{equation}

Ultimately, by individually supervising and learning vehicles of varying difficulty levels, the model's optimization direction can be intentionally adjusted to further improve the detection performance of dense and occluded vehicles.
Thus, the total loss is represented as
\begin{equation}
        L = L_{easy} + L_{hard}.
\end{equation}

\section{Experimental Results and Analysis} \label{sec:Exp}

\subsection{Data Annotation and Description} \label{sec:Data}

In this section, we label and present two multimodal vehicle detection benchmark datasets for remote sensing imagery. These datasets are distinguished from existing vehicle detection datasets by four unique features:
1) They are expressly crafted for multimodal vehicle detection in the context of large-scale events, with RGB information modal and height maps modal. Each dataset has varied resolutions and is collected using varied platforms.
2) They encompass densely packed and irregularly arranged objects, including a variety of vehicle styles as well as tents and branches.
3) They feature a wide range of occlusions, such as those caused by tents and branches.
4) They increase the complexity of vehicle detection due to the presence of distorted vehicles and the varied distribution of vehicles across large-scale areas.

The data annotation method employed utilizes the oriented bounding box (OBB) format, represented as $(x_c,y_c,w,h,\theta)$, where $(x_c,y_c)$ denotes the center coordinates, $w$ and $h$ specify the width and height of the bounding box, and $\theta$ represents the rotation angle relative to the horizontal axis of the standard bounding box. The detailed descriptions of the two multimodal datasets are as follows:

\begin{figure*}[h!]
    \begin{center}
        \includegraphics[width=1.0\textwidth]{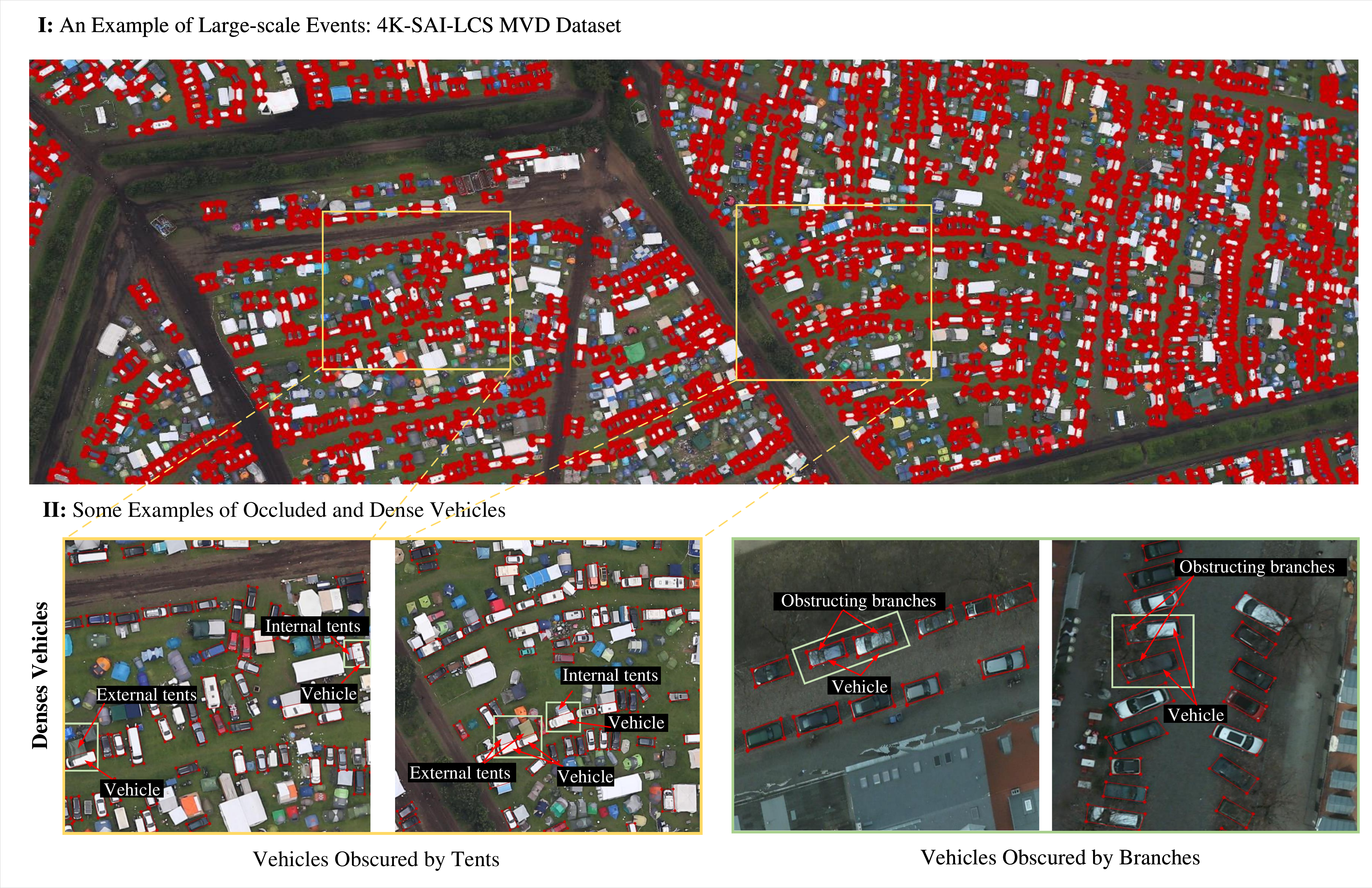}
        \caption{Visualization examples of annotated scenes and vehicles in a multimodal vehicle dataset. \textbf{I}: Sample DLR large-scale event annotated images. \textbf{II}: Vehicles under dense and occluded conditions, with density primarily due to irregularly clustered parking and occlusion caused by tents and branches.}
        \label{fig:lable}
    \end{center}
\end{figure*}

\begin{table}[!t]
\renewcommand\arraystretch{1.5}
\centering
\caption{The statistics of two multimodal object detection datasets in remote sensing. OBB is short for oriented bounding box.}
\resizebox{0.5\textwidth}{!}{
\begin{tabular}{c|c|c|c|c}
\toprule[1.5pt]
\multirow{2}{*}{Dataset} & \multicolumn{4}{c}{Attributes}\\\cline{2-5}
& Annotation &\# Categoried &\#Instances & Image resolution\\\hline
ISPRS Potsdam &OBB & \textcolor{red}{1} &4,896&$6,000\times 6,000$\\
4K-SAI-LCS &OBB & \textcolor{red}{2} &  339,111 &$5,184\times 3,456$\\ 
\bottomrule[1.5pt]
\end{tabular}
}
\label{tab:dataset}
\end{table}

\begin{figure*}[!t]
    \centering
    \subfigure[4K-SAI-LCS Dateset] {\includegraphics[width=0.495\textwidth]{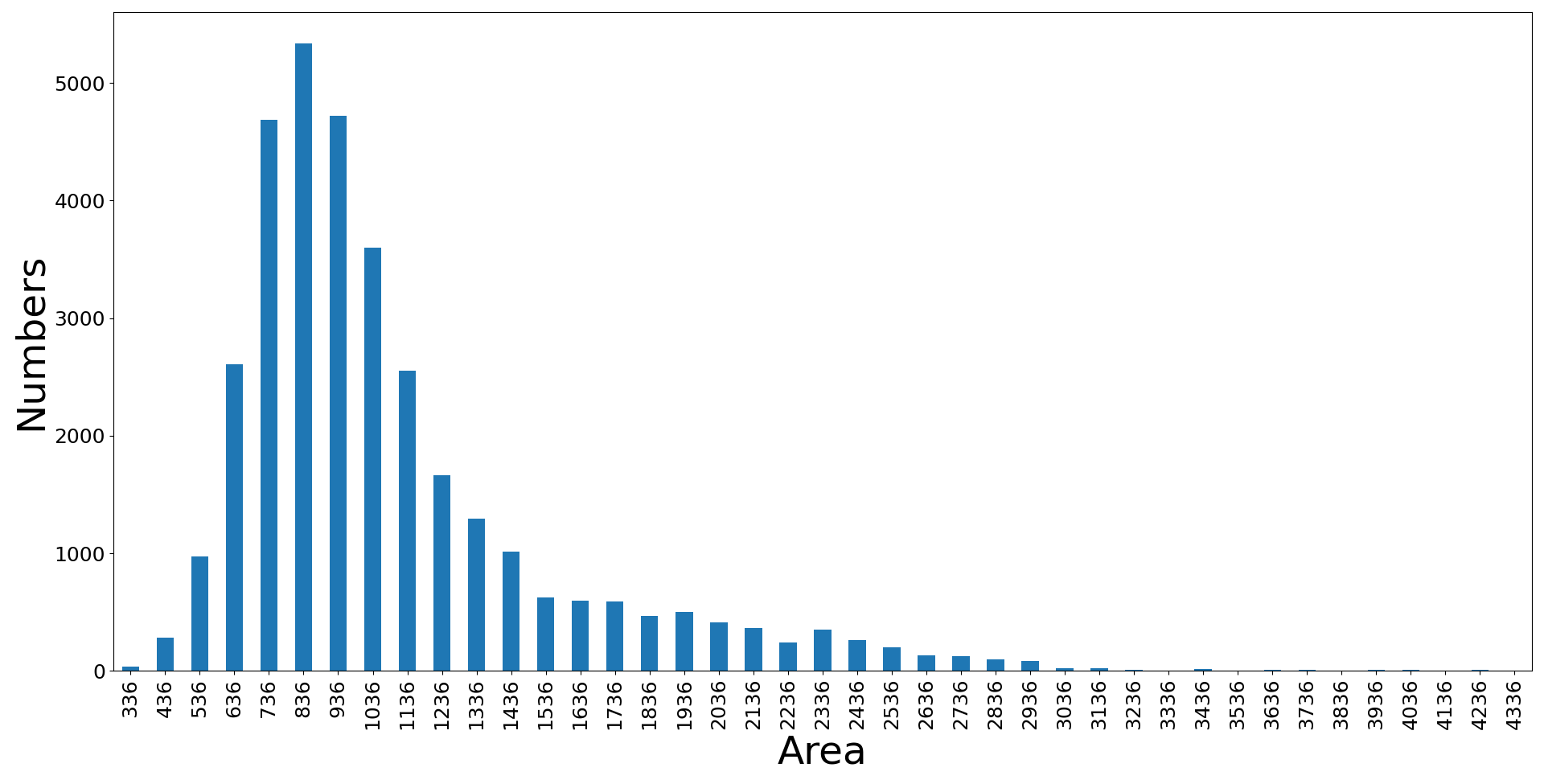}}
    \subfigure[ISPRS Potsdam Dataset]{\includegraphics[width=0.495\textwidth]{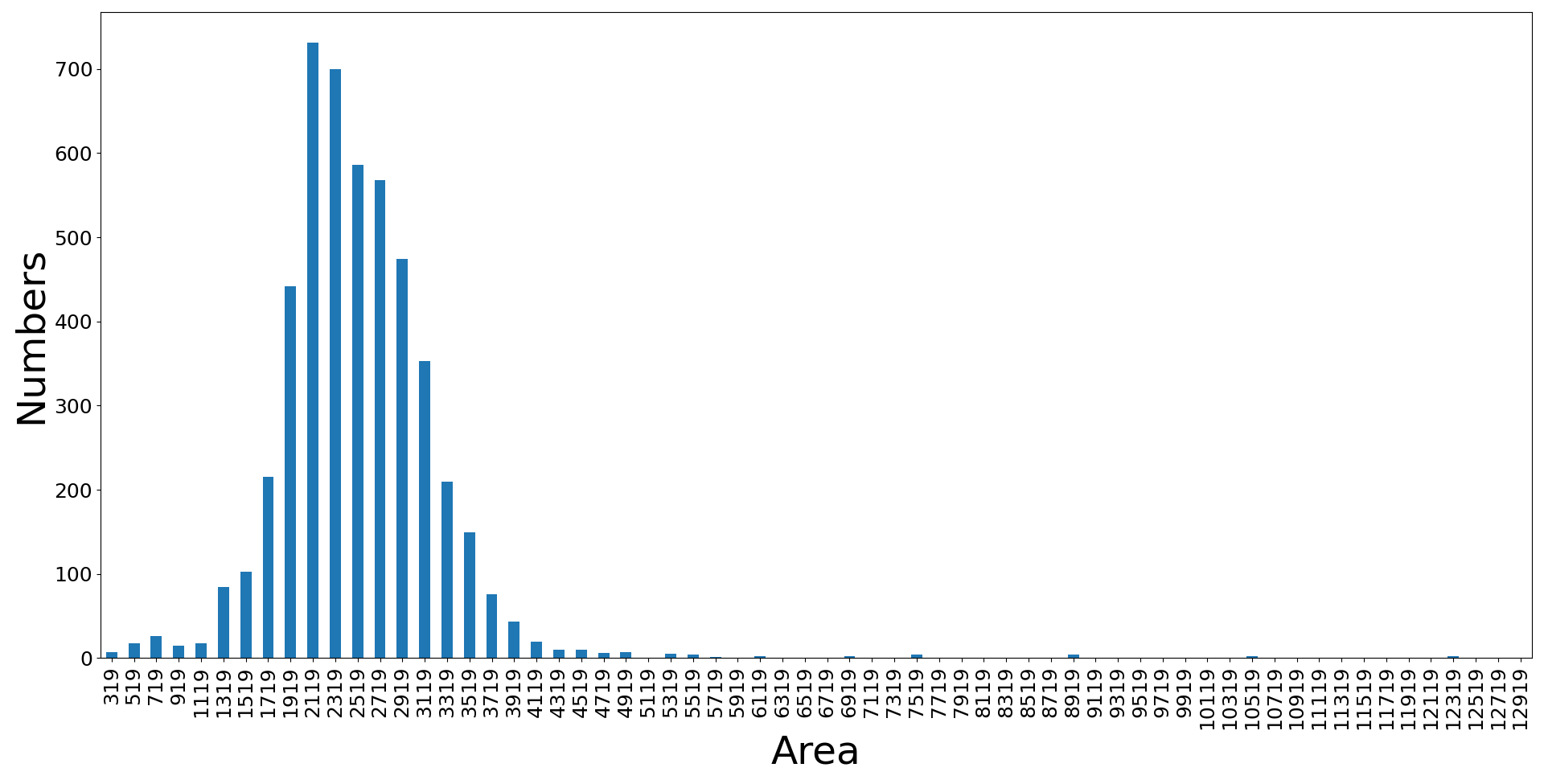}}
\caption{Statistics of vehicle instances in two multimodal vehicle detection datasets.}
\label{fig:Statistic}
\end{figure*}

2) \textit{4K-SAI-LCS MVD Dataset:}

The 4K Stereo Aerial Imagery of a Large Camping Site (4K-SAI-LCS) dataset is a subset of aerial imagery acquired from a large-scale site \cite{gstaiger20182d, wu2020vehicle}. This dataset encompasses an expanse of $1.0\times 1.5$ km. Utilizing the German Aerospace Center's advanced optical 4K camera system, a total of $114$ images were captured $54$ images with a leftward orientation and $60$ images with a rightward orientation. These images were obtained at altitudes of $600$m and $650$m above ground level, respectively. Image pre-orientation is performed by using the open source SRTM (The Shuttle Radar Topography Mission) data and measured GPS positions of the image projection centers. Precise image orientation is then accomplished by bundle adjustment using automatically extracted SIFT (Scale-Invariant Feature Transform)- tie points \cite{kurz2012low}. Afterward, the 3D point cloud is calculated using semi-global matching \cite{tian2013region, d2016improving, kempf2021oblique}. 

To preserve the original rich textures and the sharp boundaries of the vehicles in RGB images, instead of generating true orthophoto (TOP) and DSM images, in this paper, the multimodel dataset consists of the original RGB images and height maps. Unlike DSMs, height maps use the original image coordinates instead of geo-coordinates, facilitating a more direct correspondence with the RGB imagery.  
To further improve the point density, we use each test region with 4-6 overlapping images. Point clouds, created from different views, are merged and filtered. This process ensures that, after projection, there is a one-to-one relationship between each pixel in the 2D height map and its corresponding pixel in the RGB image. Both the images and height maps have ground sampling distances of $11$ cm, and each scene has a resolution of $5,184\times 3,456$.

We have annotated the 4K-SAI-LCS MVD dataset in the oriented bounding box (OBB) format using the LabelMe toolbox. This newly labeled dataset is now employed for multimodal occluded and dense vehicle detection. Fig. \ref{fig:lable}. I provide an example annotation image. The designated control zone of the festival scene encompasses a spacious parking lot and tent area. As a result, the primary objects depicted in the scene images include vehicles, tents, roads, and sanitation facilities, effectively representing the campground environment. The dataset presents significant challenges due to the dense and irregular parking arrangements of vehicles, which fall into diverse subcategories, including cars, transport vehicles, transport trailers, recreational vehicles, and camping trailers. This diversity leads to substantial intra-class variation. Moreover, the visual resemblance between vehicles and tents significantly complicates the task of vehicle detection, thereby increasing the complexity of the dataset and placing higher demands on the algorithms designed for detection. Fig. \ref{fig:lable}. II presents some examples of vehicles with varying degrees of occlusion and density.

1) \textit{ISPRS Potsdam City MVD Dataset:}\footnote{\url{http://www2.isprs.org/commissions/comm3/wg4/2d-sem-label-potsdam.html.}} The original Potsdam dataset was constructed for the ``semantic segmentation competition" by the ISPRS III/4 working group and was first published in the ISPRS 2D semantic labeling contest. Potsdam is a typical historical city and this dataset includes $38$ different regions with true orthophoto (TOP) and digital surface models (DSMs). The TOP images were captured using Trimble INPHO OrthoVista, while the DSMs, detailing the absolute elevation values for each pixel, were produced via dense image-matching techniques utilizing Trimble INPHO 5.3 software. Both images feature a ground sampling distance of $5$ cm and a resolution of $6,000\times 6,000$ pixels.

Different from existing segmentation labels, we have re-annotated the Potsdam dataset in the OBB format for images that encompass both VIS and height map data. It features vehicles located in extensive building complexes, narrow lanes, and densely populated residential zones. The designated parking areas display notable overlaps and occlusions, posing challenges in distinguishing black vehicles, particularly those obscured by foliage. Fig. \ref{fig:lable}. II presents some examples of vehicles with varying degrees of occlusion and density.

3) \textit{Dataset Statistic:}
Table \ref{tab:dataset} lists detailed instance counts for the two multimodal vehicle datasets. Given the varying scenes of image acquisition, the vehicle density in the 4K-SAI-LCS dataset is higher than in the ISPRS dataset. The 4K-SAI-LCS dataset contains over 300,000 instances, while the ISPRS dataset comprises approximately 5,000 instances. The presence of differently dense targets also poses a significant challenge for detecting densely packed vehicles. Fig. \ref{fig:Statistic} displays a curve graph comparing vehicle area to the number of vehicles in both datasets. This graph highlights the diversity and balanced distribution of vehicle objects within the two datasets. 

\subsection{Experimental Setup} \label{sec:Data}

In the experiment, data preprocessing and augmentation were applied to all images to prevent overfitting. For the 4K-SAI-LCS dataset, input images were cropped to three sizes: $640\times 640$, $800\times 800$, and $1,024\times 1,024$ pixels, each with a $200$-pixel overlap to maximize object information capture. The training and testing sets were equally divided, maintaining a $1:1$ ratio. To balance the volumes of the two multimodal datasets, $12$ original images from the ISPRS Potsdam City Dataset, featuring aligned VIS and DSM scenarios, were selected. These input images were cropped to $800\times 800$ pixels, with a $400$-pixel overlap for optimal object information capture. The ratio of the training set to the testing set for the ISPRS dataset was adjusted to $784:392$ to align with the experimental requirements."

The initial and final learning rates are set at $1.5\times 10^{-4}$ and $1\times 10^{-6}$, respectively. We employ the Stochastic Gradient Descent (SGD) optimization strategy, with a weight decay of $5\times 10^{-4}$ and momentum of $0.9$. The training process is designed to run for a maximum of $200$ epochs, with a confidence threshold set at $0.2$ and a Non-Maximum Suppression (NMS) threshold of $0.45$. Given the distinct information content of height maps and RGB data, we utilize different backbone networks for each data stream: ResNet18 for height maps and Darknet53 for RGB data. The proposed MuDet architecture is implemented using the PyTorch framework on an NVIDIA GeForce RTX 3090 GPU.

\begin{figure*}[!t]
    \begin{center}
        \includegraphics[width=1.0\textwidth]{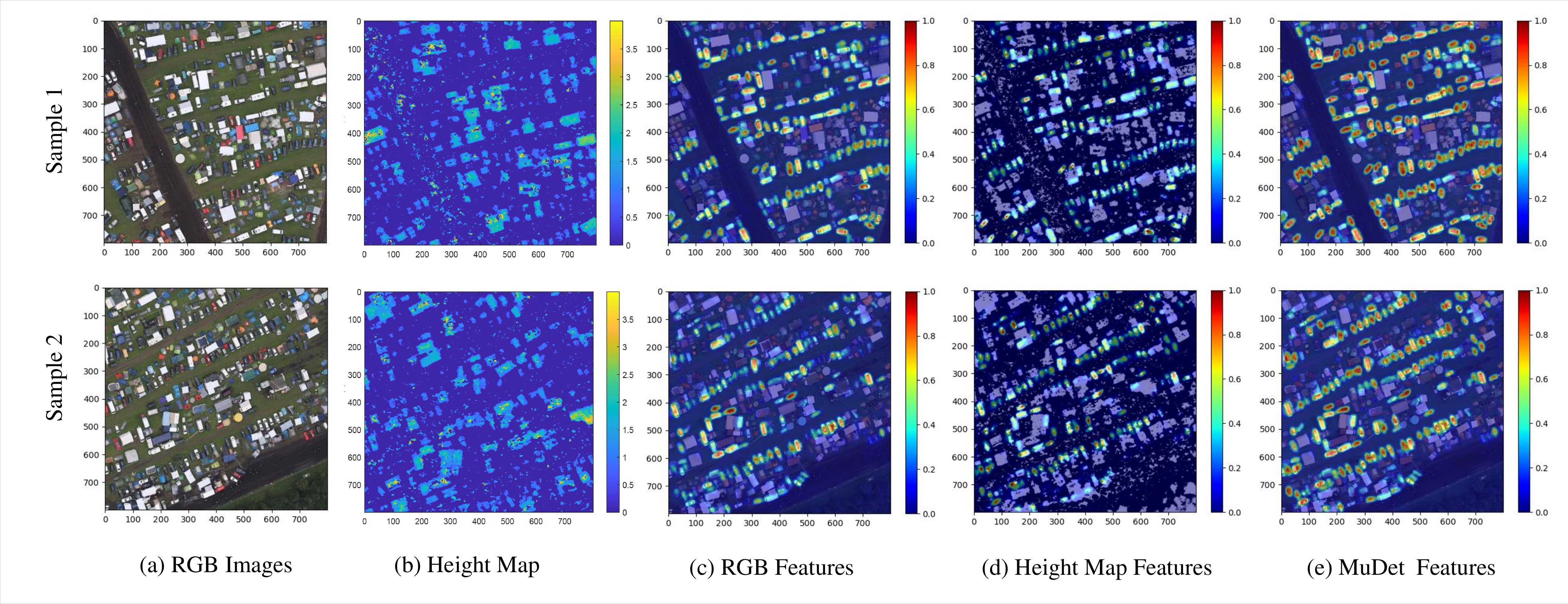}
        \caption{Vehicle separation visualization results by modality increment by successively using RGB images, Height map images, RGB attention features, Height map features, and our proposed MuDet on 4K-SAI-LCS dataset.}
        \label{fig:fea}
    \end{center}
\end{figure*}

\subsection{Evaluation Metrics} 
Three common object detection evaluation criteria are utilized for quantitative analysis, including \textit{Precision (P)}, \textit{Recall (R)}, and \textit{Average Precision ($AP$)}. 
\begin{equation}
\begin{aligned}
       P &= \frac{TP}{TP+FP}\\
       R &=\frac{TP}{TP+FN}, \\
\end{aligned}
\end{equation}
where $TP$, $FP$, and $FN$ represents true positive, false positive objects, and false negative objects, respectively. Generally, higher values of these metrics indicate superior detection performance.

\begin{table}[!t]
\renewcommand\arraystretch{1.5}
\centering
\small
\caption{The contribution of different modalities. The best results are shown in bold.}
\resizebox{0.5\textwidth}{!}{
\begin{tabular}{c|c|c||c|c}
\toprule[1.5pt]
\multirow{2}{*}{Modality} &\multirow{2}{*}{Attention} & \multirow{2}{*}{Backbone} &ISPRS Potsdam & 4K-SAI-LCS\\\cline{4-5}
 && & AP0.5(\%) & AP0.5(\%)\\\hline
RGB & -&Darknet53 & 90.03 &86.99\\
Height map & -&ResNet18 & 13.13 &  27.04\\ 
RGB &Self Attention &Darknet53 & 91.47 &90.02\\
Height map &Self Attention &ResNet18 & 11.94 &  27.53\\ 
MuDet(v3) &Mul-Lea &Darknet53/ResNet18 & 93.63 &  92.57\\ 
MuDet(GGHL) &Mul-Lea &Darknet53/ResNet18 & 94.58 &  94.19\\ 
MuDet(v8) &Mul-Lea &CSPDarknet53/ResNet18 & \bf 94.92 &  \bf 95.07\\
\bottomrule[1.5pt]
\end{tabular}
}
\label{tab:modality}
\end{table}
\begin{table}[!t]
\renewcommand\arraystretch{1.5}
\centering
\small
\caption{The contribution of the fusion method. The best results are shown in bold.}
\resizebox{0.5\textwidth}{!}{
\begin{tabular}{c|c||c|c}
\toprule[1.5pt]
\multirow{2}{*}{Fusion method} & \multirow{2}{*}{Network} &ISPRS Potsdam& 4K-SAI-LCS\\\cline{3-4}
 & & AP0.5(\%) & AP0.5(\%)\\\hline
Image-level  &Darknet53 & 93.12 & 87.89\\
Feature-level  &Darknet53/ResNet18 & 94.05&  89.27\\ 
Uni-Enh+Mul-Lea  &Darknet53/ResNet18 & 94.37 & 91.11\\
Feature-level+He-Dis  &Darknet53/ResNet18 & 94.28 & 90.64\\
MuDet(v3)  &Darknet53/ResNet18 & 93.63 &  92.57\\ 
MuDet(GGHL) &Darknet53/ResNet18 & 94.58 &  94.19\\ 
MuDet(v8) &CSPDarknet53/ResNet18 & \bf 94.92 &  \bf 95.07 \\
\bottomrule[1.5pt]
\end{tabular}
}
\label{tab:fusion}
\end{table}

\textit{$AP$} is a global indicator, enabling fair comparison across different detection methods. In our experiments, $AP0.5$ refers to the Average Precision (AP) calculated at an Intersection over Union (IoU) threshold of 0.5.

\begin{equation}
    AP= \sum_{k=1}^{n}P(k)\Delta R(k),
\end{equation}
where $k$ represents the threshold. $P(k)$ denotes the precision at the $k$-th threshold. is the precision at the $k$-th threshold. $\Delta R(k)=R(k)-R(k-1)$ calculates the change in recall between consecutive $k-1$-th and $k$-th thresholds.

\begin{figure*}[!t]
    \begin{center}
        \includegraphics[width=1.0\textwidth]{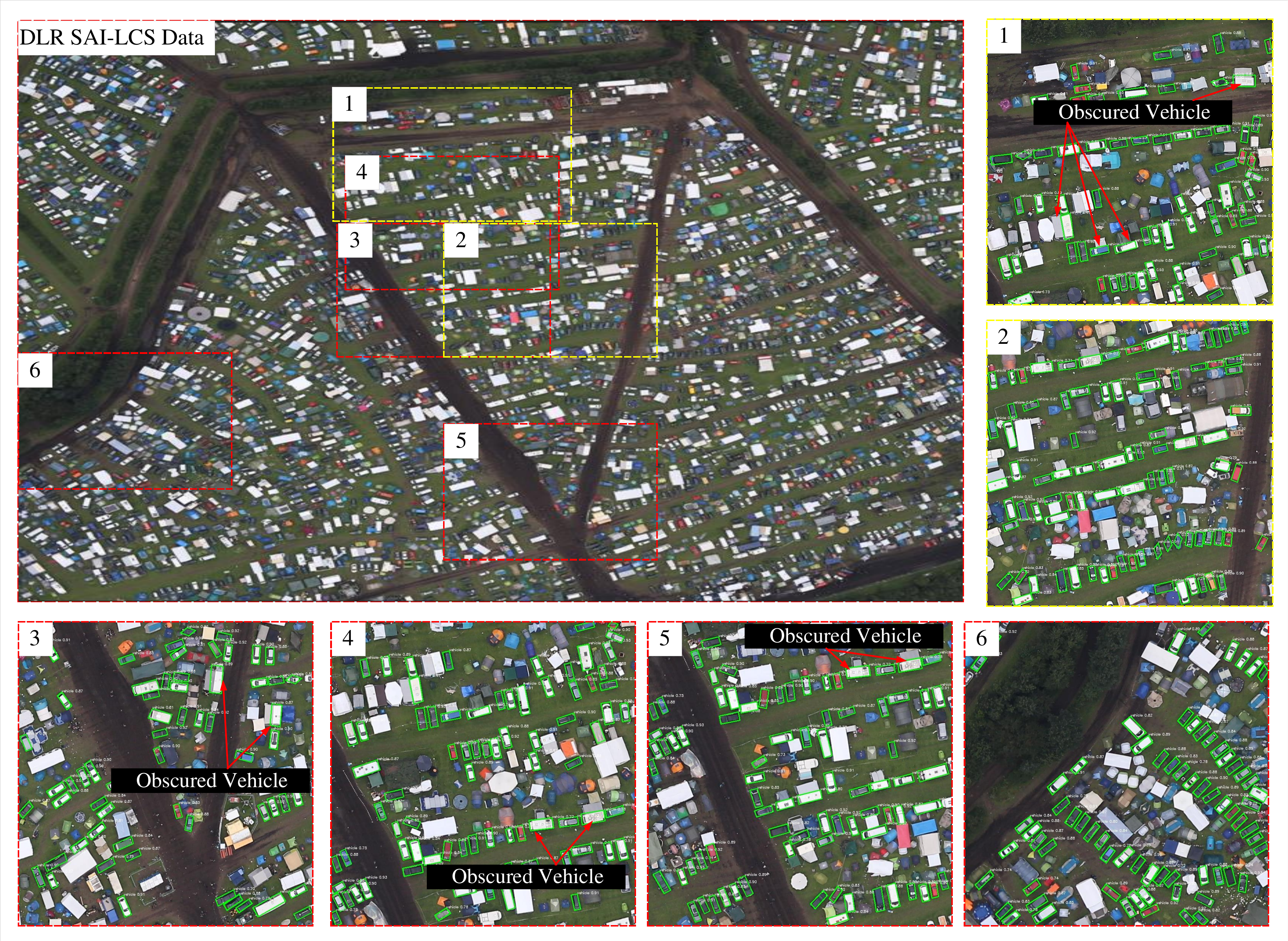}
        \caption{The visualization results of MuDet on selected sample images from the 4K-SAI-LCS dataset. The vehicle density and occlusion are predominantly due to the restricted event area, irregular parking within the area, and coverage by tents, respectively. } 
        \label{fig:DLR}
    \end{center}
\end{figure*}

\begin{table*}[!t]
\renewcommand\arraystretch{1.5}
\centering
\caption{Average Precision (AP) results in comparisons for the 4K-SAI-LCS dataset. The best results are highlighted in bold.}
\resizebox{1.0\textwidth}{!}{
\begin{tabular}{c|cccccc|ccc}
\toprule[1.0pt]

&\multirow{2}{*}{YOLOv3\cite{redmonYOLOv3IncrementalImprovement2018} }& \multirow{2}{*}{RetinaNet\cite{lin2017focal}} &\multirow{2}{*}{FCOS\cite{tian2019fcos}} &\multirow{2}{*}{GGHL\cite{huang2022general}} & \multirow{2}{*}{YOLOv8 \cite{yolov8}} &\multirow{2}{*}{RepPoints\cite{yang2019reppoints}} & \multicolumn{3}{c}{MuDet} \\ \cline{8-10}
&&&&&&&V3-based&GGHL-based& V8-based \\ \hline
Unimodal&81.75& 83.44 &84.81& 86.99 & 87.26 & 87.98  & -& -& -\\ \hline
Multimodal&82.91& 84.78 &85.91 & 89.27 & 88.53 &90.71 & 92.57 & 94.19 & \bf{95.07} \\
\bottomrule[1.0pt]
\end{tabular}
}
\label{tab:DLR}
\end{table*}

\subsection{Comparison with state-of-the-art MVD models } 
In the experiment, six commonly used methods for vehicle detection in multimodal remote sensing (RS) images were selected for both quantitative and qualitative comparisons. These methods include You only look once (YOLOv3)\cite{redmonYOLOv3IncrementalImprovement2018}, RetinaNet\cite{lin2017focal}, Fully Convolutional One-Stage object detector (FCOS)\cite{tian2019fcos}, General Gaussian Heatmap Label Assignment (GGHL)\cite{huang2022general}, Representative Points (RepPoints)\cite{yang2019reppoints}, and YOLOv8 \cite{yolov8}. Darknet53 was employed as the backbone network across all methods, complemented by a multi-scale feature pyramid network (FPN) for upsampling and fusion to ensure a balanced comparison.

\subsection{Ablation Study} 
In this section, we evaluated the contribution of the proposed MuDet through two ablation analysis experiments. Specifically, 1) the contribution of the modality increment; 2) the contribution of the fusion increment.

\subsubsection{Contribution of the Modality Increment.}
Table \ref{tab:modality} quantifies the improvement in vehicle detection performance by the incremental addition of modalities. Note that RGB images offer rich color and texture information, whereas height maps supply solely elevation data. Employing distinct backbones for each modality, Darknet53/CSPDarknet53 for RGB and ResNet18 for height map, enables the extraction of the most valuable features from each modality while avoiding overfitting. For unimodal, height information alone proves insufficient for distinguishing vehicles without the complementary support of RGB data. The performance enhancement achieved by integrating self-attention into each modality falls short of the improvements in three variants of MuDet (the YOLOv3-based MuDet (MuDet(v3)), the DDHL-based MuDet (MuDet(GGHL)), and the YOLOv8-based MuDet (MuDet(v8)), with the largest gap in performance exceeding 5\%. Fig. \ref{fig:fea} shows the visualization of vehicle separation results through incremental modality utilization, including RGB images, height map images, RGB + self attention, height map + self attention, and yolov8-based MuDet, when applied to the 4K-SAI-LCS dataset. MCo-Net outperforms other methods in separating densely packed vehicles.

\subsubsection{Contribution of the Fusion Increment.}
Table. \ref{tab:fusion} quantifies the improvement in vehicle detection performance by the various fusion strategies, including \textit{image-level fusion}, \textit{feature-level fusion}, \textit{Uni-Enh+Mul-Lea}, \textit{feature-level+He-Dis}, and three types of MuDet. Overall, both image-level fusion and feature-level fusion result in the poorest detection performance. The best MuDet (v8) achieves an approximate 6\% improvement in AP for the 4K-SAI-LCS dataset and a 1\% improvement in AP for the ISPRS Potsdam dataset. In contrast, the \textit{Feature-Level+He-Di} method, which utilizes feature fusion, demonstrates minimal improvement due to its lack of consideration for the interplay among multimodal features. Note that the vehicle categories within the ISPRS Potsdam dataset are relatively uniform, and their density is low. Consequently, the improvement offered by the feature-level+He-Dis module is somewhat constrained.
 
\subsection{Results and Analysis on the 4K-SAI-LCS Data} \label{sec:Exp_Dis}

\begin{figure*}[!t]
    \centering
    \subfigure[4K-SAI-LCS Dateset] {\includegraphics[width=0.45\textwidth]{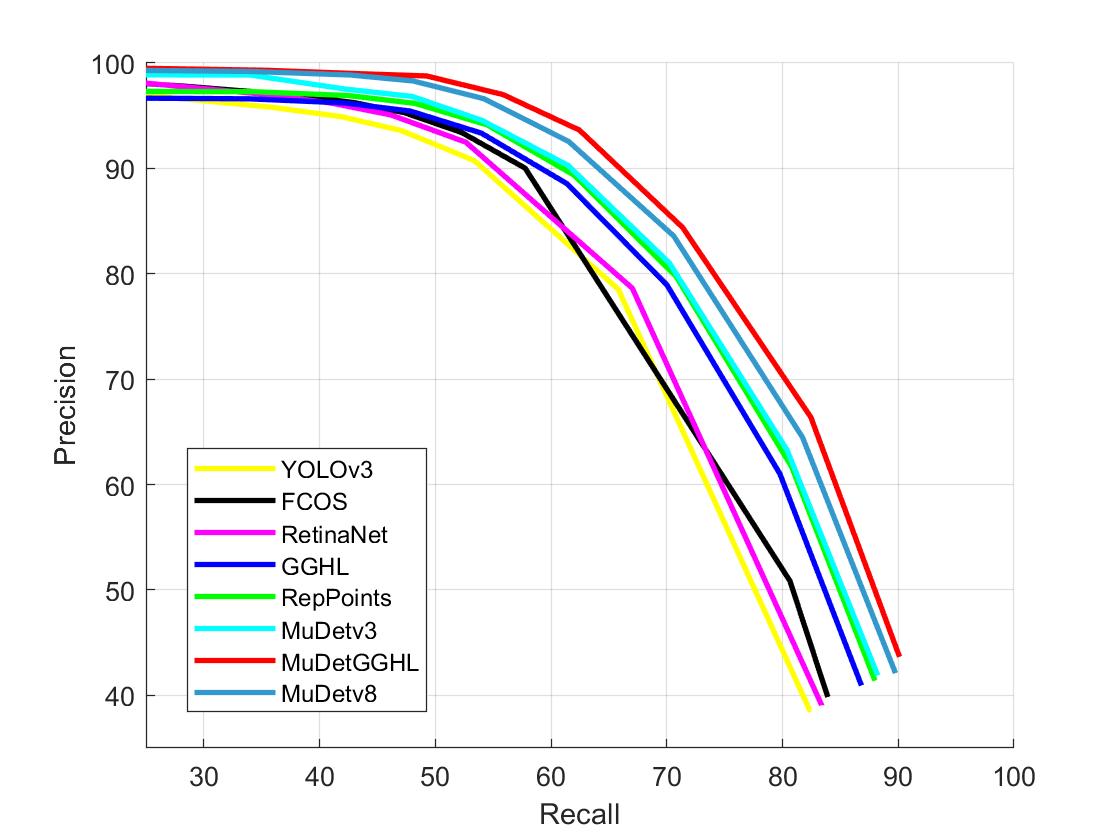}}
    \subfigure[ISPRS Potsdam Dataset]{\includegraphics[width=0.45\textwidth]{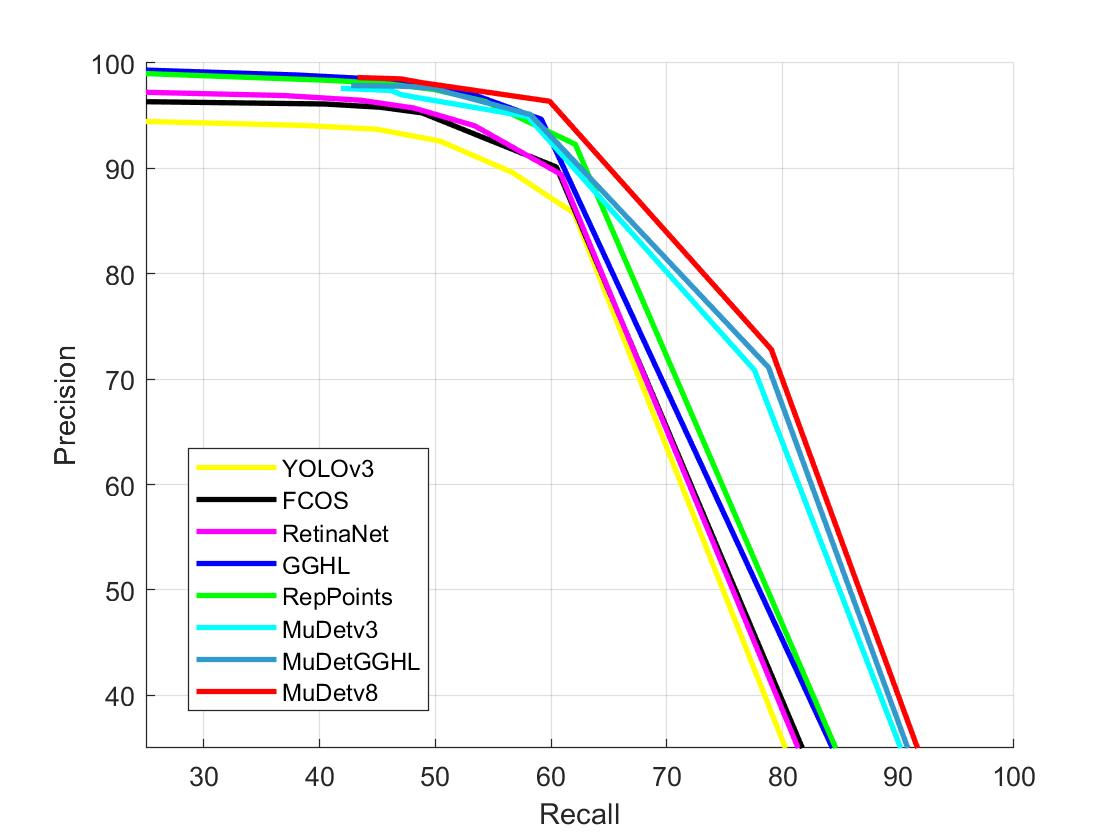}}
\caption{PR Curves of the proposed MuDet in comparison with six methods.}
\label{PR Curve}
\end{figure*}

Table. \ref{tab:DLR} lists the quantitative detection accuracy of the six methods for the 4K-SAI-LCS dataset in terms of AP. Overall, multimodal data significantly outperforms single modality data in terms of AP value. Specifically, RetinaNet, due to its focal loss design, reduces the weight of easily detected objects, achieving a detection performance approximately 2\% higher than YOLOv3. FCOS, GGHL, and YOLOv8, all anchor-free methods, show distinct advantages over anchor-based methods. FCOS enhances detection precision by predicting object presence at each pixel, thus avoiding the complex anchor matching process. GGHL employs a Gaussian heatmap distribution technique to improve the learning capability of objects of different sizes in various positions. YOLOv8 has been redesigned with a regression-based loss function and sample matching strategies, among other modules, offering significant improvements in both speed and accuracy, approximately 6\% higher than YOLOv3. Unlike traditional anchor-based methods, Reppoints models each object's unique shape and contour without relying on predefined anchor box sizes or ratios, achieving optimal performance in single-modality object detection that may be influenced by background or semantically irrelevant foreground information. The proposed MuDet increases detection accuracy across different network backbones, including YOLOv3, GGHL, and YOLOv8, by over 5\% compared to their corresponding single modality. MuDet introduces single-modality feature enhancement, multimodal cross-fusion, and a strategy for balancing samples of varying difficulty, effectively segregating vehicles, and eliminating the impact of background information. 

Fig. \ref{fig:DLR} shows the visualization results of the MuDet on selected example images, demonstrating effective separability for dense and occluded vehicles, such as RVs with open doors and tents mounted on cars. However, a small fraction of white recreational vehicles (RVs) were missed due to the absence of open-door samples in the labeled 4K-SAI-LCS vehicle dataset. Additionally, we have to admit that the 4K-SAI-LCS dataset presents significant challenges for vehicle detection. Notably, some vehicles are difficult to distinguish, with their similarity to tents sometimes exceeding 80\%.

Given the substantial impact that varying confidence thresholds can have on model performance, it is crucial to analyze their effects. Fig. \ref{PR Curve} (a) illustrates the Precision-Recall (PR) curve for all comparative methods and the proposed MuDet across different thresholds. It can be shown that the proposed method not only achieves the highest precision for a given level of recall but also has the largest area under the PR curve, reaffirming the efficacy and superiority of MuDet. However, the recall metric still warrants enhancement. This likely stems from the significant intra-class variance and subtle inter-class variance. The challenge in achieving precise fine-grained separation becomes apparent when merging all patterned vehicles into a broad "vehicle" category, especially when distinct types, such as cars and trucks, are amalgamated into a single category, hindering precise differentiation. 

\begin{figure*}[!t]
    \begin{center}
        \includegraphics[width=1.0\textwidth]{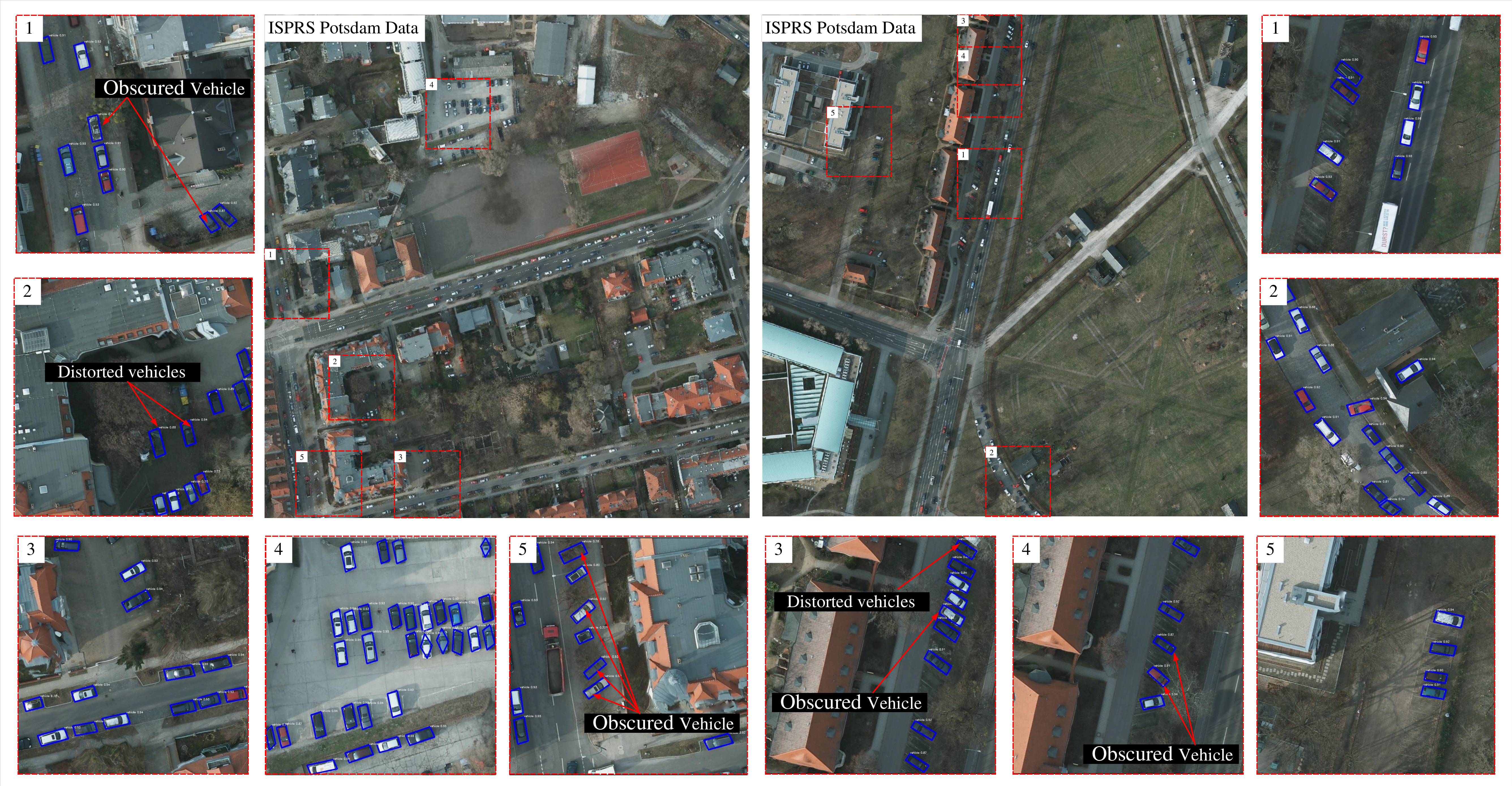}
        \caption{The visualization results of MuDet on selected sample images from the ISPRS Potsdam city dataset. The vehicle density, occlusion, and distortion are primarily attributed to parking lots, branches, and the image capture process, respectively.}
        \label{fig:Pot}
    \end{center}
\end{figure*}

\begin{table*}[!t]
\renewcommand\arraystretch{1.5}
\centering
\caption{Average Precision (AP) results in comparisons for the ISPRS Potsdam City dataset. The best results are highlighted in bold.}
\resizebox{1.0\textwidth}{!}{
\begin{tabular}{c|cccccc|ccc}
\toprule[1.0pt]
&\multirow{2}{*}{YOLOv3\cite{redmonYOLOv3IncrementalImprovement2018} }& \multirow{2}{*}{RetinaNet\cite{lin2017focal}} &\multirow{2}{*}{FCOS\cite{tian2019fcos}} &\multirow{2}{*}{GGHL\cite{huang2022general}} & \multirow{2}{*}{YOLOv8 \cite{yolov8}} &\multirow{2}{*}{RepPoints\cite{yang2019reppoints}} & \multicolumn{3}{c}{MuDet} \\ \cline{8-10}
&&&&&&&V3-based&GGHL-based& V8-based \\ \hline
Unimodal&85.66 &88.10 &88.80  & 90.03 &90.39 & 91.46  & -& -& - \\ \hline
Multimodal&87.42 &88.42 &89.08  & 94.05 &91.70 & 93.65 & 93.63 & 94.58 & \bf{94.92} \\
\bottomrule[1.0pt]
\end{tabular}
}
\label{tab:Pot}
\end{table*}

\subsection{Results and Analysis on the ISPRS Potsdam City Data} \label{sec:Exp_Dis}
Table \ref{tab:Pot} lists a quantitative performance analysis of the ISPRS Potsdam city dataset, while Fig. \ref{fig:Pot} shows the visualization results for sample images using our methods. 

Overall, the detection performance on the Potsdam city dataset is consistent with that on the 4K-SAI-LCS dataset, further demonstrating that MCo-Net improves the detection performance of dense and occluded vehicle objects. MuDet achieves a significant improvement over unimodal YOLOv3 and multimodal YOLOv3, with increases of 10\% and 8\%, respectively. Relying solely on single-modality data raises the likelihood of inaccurate detections or missed vehicles. Especially dark vehicles are obscured by tree branches or closely match the color of the branches. Furthermore, compared to competitive multimodal methods based on YOLOv8 and Reppoint, MuDet achieves improvements of up to 3\% and 1.5\%, respectively, further verifying the method's robustness.

Fig. \ref{PR Curve} (b) shows the PR curve for all comparison methods and the proposed MuDet under dynamic thresholds. MuDet obtains the largest area under the PR curve, similar to the performance on the 4K-SAI-LCS dataset, further demonstrating MuDet's effectiveness and superiority. However, at equivalent levels of accuracy, this dataset exhibits a lower recall rate compared to the 4K-SAI-LCS dataset. This discrepancy is likely attributed to varying occlusion levels from tree branches and vehicle deformation, which not only diminishes the contrast between background and foreground but also amplifies the intra-class variation of vehicle targets, leading to a reduced recall rate.

Fig. \ref{fig:Pot} shows some detection results on the ISPRS Potsdam city dataset. While this dataset has a lower vehicle density compared to the 4K-SAI-LCS dataset, it also features other challenges, such as black vehicles obscured by tree branches and vehicles with deformations. For these vehicles, MuDet achieved commendable detection results. However, there remains a considerable opportunity for further refinement, particularly in improving the detection of vehicles with significant deformations or those extensively occluded by tree branches.
		
\section{Conclusion}\label{sec:clu}
In this article, we initially develop two multimodal datasets for dense and occluded vehicle detection in large-scale scenarios, employing both RGB and height map modalities. Subsequently, we propose a multimodal collaboration network, termed MuDet, for dense and occluded vehicle detection in large-scale events. MuDet is designed to fully exploit unimodal enhanced features, multimodal cross-features, and patterns that distinguish between hard and easy vehicle detection. Leveraging the integrated data from RGB and height maps, MuDet excels in differentiating vehicles based on color, identifying vehicles with similar colors and textures in crowded scenes through their unique height values, and detecting occluded vehicles, thereby enhancing its utility in complex environments. Extensive experiments conducted on two newly constructed and labeled image datasets demonstrate MuDet's superiority in MVD compared to commonly employed detection methods.

However, MuDet currently does not adapt well to two multimodal datasets with significant distributional variance. Therefore, future work will focus on exploring domain adaptation techniques in multimodal contexts to improve the model's ability to generalize effectively across various domains.

\section*{Acknowledgement}

\bibliographystyle{ieeetr}
\bibliography{wx_ref}

\begin{IEEEbiography}[{\includegraphics[width=1.15in,height=1.25in,clip,keepaspectratio]{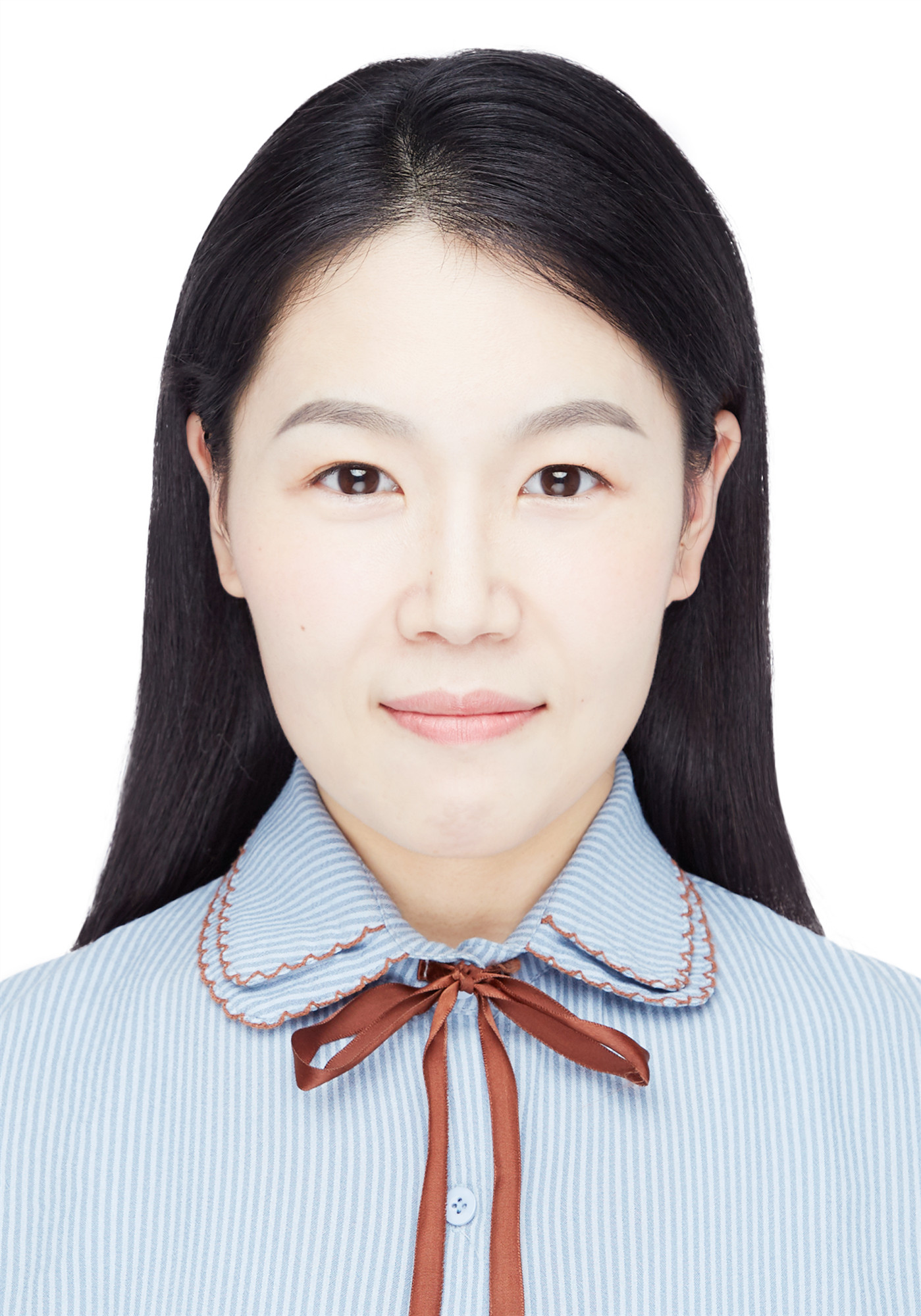}}]{Xin Wu} (Senior Member, IEEE) received the Ph.D. degree from the School of Information and Electronics, Beijing Institute of Technology (BIT), Beijing, China, in 2020. She is currently an Assistant Professor at the School of Computer Science, Beijing University of Posts and Telecommunications (BUPT), Beijing, China. Her research interests include deep learning, remote sensing, object detection, and multimodal intelligent perception.

Dr. Wu is a Topical Associate Editor of the IEEE Transactions on Geoscience and Remote Sensing (TGRS). She was a recipient of the Best Reviewer Award of the IEEE TGRS in 2023 and the IEEE JSTARS in 2022 as well as the Jose Bioucas Dias award for recognizing the outstanding paper at the Workshop on Hyperspectral Imaging and Signal Processing: Evolution in Remote Sensing (WHISPERS) in 2021. She is also a Leading Guest Editor of the IEEE JSTARS and Remote Sensing.
\end{IEEEbiography}

 \vskip -2\baselineskip plus -1fil

 \begin{IEEEbiography}[{\includegraphics[width=1in,height=1.25in,clip,keepaspectratio]{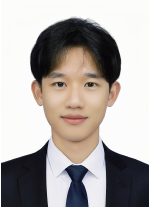}}]{Zhanchao Huang}(Member, IEEE) received the Ph.D. degree from the School of Information and Electronics, Beijing Institute of Technology (BIT), Beijing, China, in 2023. He is currently an Assistant Professor at The Academy of Digital China (ADC), Fuzhou University (FZU), Fuzhou 350108, China. His research interests include object detection and remote sensing image interpretation.
 \end{IEEEbiography}
 
\vskip -2\baselineskip plus -1fil

\begin{IEEEbiography}[{\includegraphics[width=1in,height=1.25in,clip,keepaspectratio]{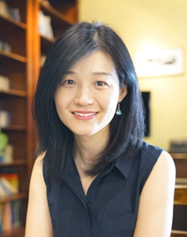}}]{Li Wang} (Senior Member, IEEE) received the Ph.D. degree from the Beijing University of Posts and Telecommunications (BUPT), Beijing, China, in 2009. She is currently a Full Professor with the School of Computer Science, National Pilot Software Engineering School, BUPT, where she is also an Associate Dean and the Head of the High Performance Computing and Networking Laboratory.
She is also a Member of the Key Laboratory of the Universal Wireless Communications, Ministry of Education, China. She is also a rotating director of the Key Laboratory of Application Innovation in Emergency Command Communication Technology, Ministry of Emergency Management, China. 
She also held Visiting Positions with the School of Electrical and Computer Engineering, Georgia Tech, Atlanta, GA, USA, from December 2013 to January 2015, and with the Department of Signals and Systems, Chalmers University of Technology, Gothenburg, Sweden, from August to November 2015 and July to August 2018. 
She has authored or coauthored almost 70 journal papers and four books. 
Her research interests include wireless communications, distributed networking and storage, vehicular communications, social networks, and edge AI. 
She currently serves on the Editorial Boards for \emph{IEEE Transactions on Vehicular Technology}, \emph{IEEE Transactions on Cognitive Communications and Networking}, \emph{IEEE Internet of Things Journal}, and \emph{China Communications}. 
She was an Associate Editor for \emph{IEEE Transactions on Green Communications and Networking}, the Symposium Chair of IEEE ICC 2019 on Cognitive Radio and Networks Symposium, and a Tutorial Chair of IEEE VTC 2019. 
She also is the chair of the Special Interest Group (SIG) on Sensing, Communications, Caching, and Computing (C3) in Cognitive Networks for the IEEE Technical Committee on Cognitive Networks. 
She was the Vice Chair of the Meetings and Conference Committee (MCC) for the IEEE Communication Society (ComSoc) Asia Pacific Board (APB) for the term of 2020–2021. 
She was the recipient of the 2013 Beijing Young Elite Faculty for Higher Education Award, best paper awards from several IEEE conferences, IEEE ICCC 2017, IEEE GLOBECOM 2018, and IEEE WCSP 2019. 
She was also the recipient of the Beijing Technology Rising Star Award in 2018. 
She has served on the TPC of multiple IEEE conferences, including IEEE Infocom, Globecom, International Conference on Communications, IEEE Wireless Communications and Networking Conference, and IEEE Vehicular Technology Conference in recent years.
 \end{IEEEbiography}

\vskip -2\baselineskip plus -1fil

\begin{IEEEbiography}[{\includegraphics[width=1in,height=1.25in,clip,keepaspectratio]{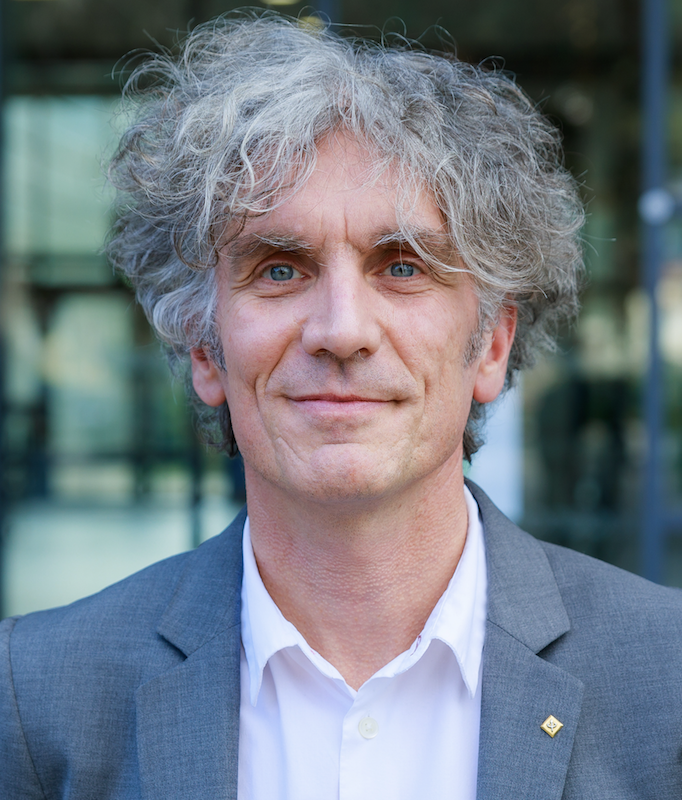}}]{Jocelyn Chanussot}
(IEEE Fellow) received the M.Sc. degree in electrical engineering from the Grenoble Institute of Technology (Grenoble INP), Grenoble, France, in 1995, and the Ph.D. degree from the Université de Savoie, Annecy, France, in 1998. From 1999 to 2023, he has been with Grenoble INP, where he was a Professor of signal and image processing. He is currently a Research Director with INRIA, Grenoble. His research interests include image analysis, hyperspectral remote sensing, data fusion, machine learning, and artificial intelligence. He has been a visiting scholar at Stanford University (USA), KTH (Sweden), and NUS (Singapore). Since 2013, he has been an Adjunct Professor at the University of Iceland. In 2015-2017, he was a visiting professor at the University of California, Los Angeles (UCLA).  He holds the AXA chair in remote sensing and is an Adjunct professor at the Chinese Academy of Sciences, Aerospace Information Research Institute, Beijing, China.

Dr. Chanussot is the founding President of the IEEE Geoscience and Remote Sensing French chapter (2007-2010) which received the 2010 IEEE GRSS Chapter Excellence Award. He was the Vice-President of the IEEE Geoscience and Remote Sensing Society, in charge of meetings and symposia (2017-2019). He is an Associate Editor for the IEEE Transactions on Geoscience and Remote Sensing, the IEEE Transactions on Image Processing, and the Proceedings of the IEEE. He was the Editor-in-Chief of the IEEE Journal of Selected Topics in Applied Earth Observations and Remote Sensing (2011-2015). In 2014 he served as a Guest Editor for the IEEE Signal Processing Magazine. He is a Fellow of the IEEE, an ELLIS Fellow, a Fellow of AAIA, a member of the Institut Universitaire de France (2012-2017), and a Highly Cited Researcher (Clarivate Analytics/Thomson Reuters, since 2018).
\end{IEEEbiography}

\vskip -2\baselineskip plus -1fil

\begin{IEEEbiography}
[{\includegraphics[width=1in,height=1.25in,clip,keepaspectratio]{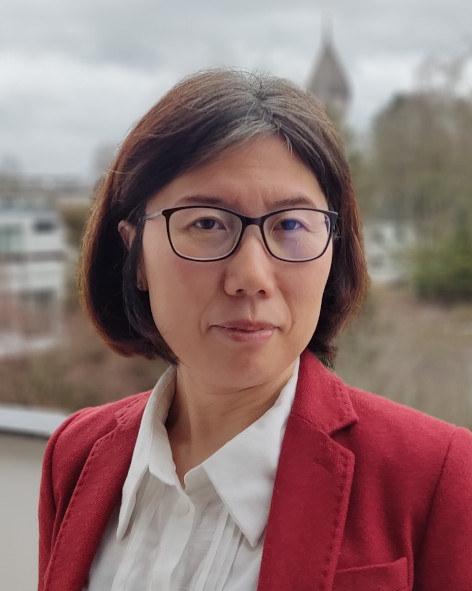}}] {Jiaojiao Tian}(M'19--SM'21) received her B.S. degree in geoinformation systems from the China University of Geoscience, Beijing, in 2006, her M. Eng. degree in cartography and geoinformation at the Chinese Academy of Surveying and Mapping, Beijing, in 2009, and her Ph.D. degree in mathematics and computer science from Osnabrück University, Germany, in 2013. Since 2009, she has been with the Photogrammetry and Image Analysis Department, Remote Sensing Technology Institute, German Aerospace Center, Wessling, Germany, where she is currently head of the 3D and Modeling Group. In 2011, she was a guest scientist with the Institute of Photogrammetry and Remote Sensing, ETH Zürich, Switzerland. She serves as a co-chair of the ISPRS Commision WG I/8: Multi-sensor Modelling and Cross-modality Fusion. She is a member of the editorial board of the ISPRS Journal of Photogrammetry and Remote Sensing and of the International Journal of Image and Data Fusion. 

Her research interests include 3D change detection, digital surface model (DSM) generation, 3D point cloud semantic segmentation, object extraction, and DSM-assisted building reconstruction, forest monitoring and classification.
\end{IEEEbiography}

\end{document}